\documentclass[conference]{IEEEtran}
\IEEEoverridecommandlockouts
\usepackage{cite}
\usepackage{comment}
\usepackage{hyperref}
\usepackage{amsmath,amssymb,amsfonts}
\usepackage{algorithmic}
\usepackage{graphicx}
\usepackage{textcomp}
\usepackage{xcolor}
\usepackage{dirtree}
\usepackage{soul}
\usepackage{pifont}
\usepackage{multirow}
\usepackage{color, colortbl}	
\definecolor{LightCyan}{rgb}{0.88,1,1}
\newcommand{\yes}{\ding{51}}
\newcommand{\no}{\ding{55}}
\usepackage[a4paper, total={184mm,239mm}]{geometry}
\def\BibTeX{{\rm B\kern-.05em{\sc i\kern-.025em b}\kern-.08em
    T\kern-.1667em\lower.7ex\hbox{E}\kern-.125emX}}

\renewcommand{\baselinestretch}{0.975}

\usepackage{tikz}
\usepackage{textcomp}
\usepackage[doipre={DOI:~}]{uri}
\usepackage{lipsum}
\newcommand\copyrighttext{%
  \footnotesize \textcopyright 2025 IEEE. Personal use of this material is permitted.  Permission from IEEE must be obtained for all other uses, in any current or future media, including reprinting/republishing this material for advertising or promotional purposes, creating new collective works, for resale or redistribution to servers or lists, or reuse of any copyrighted component of this work in other works.
 
  Accepted for publications at FDL 2025 Forum on specification \& Design Languages.}
\newcommand{\copyrightnotice}{%
\begin{tikzpicture}[remember picture,overlay]
\node[anchor=south,yshift=10pt] at (current page.south) {\fbox{\parbox{\dimexpr\textwidth-\fboxsep-\fboxrule\relax}{\copyrighttext}}};
\end{tikzpicture}%
}
\begin{document}
\bstctlcite{IEEEexample:BSTcontrol}
\title{Automatic integration of SystemC in the FMI standard for Software-defined Vehicle design
}

\author{\IEEEauthorblockN{Giovanni Pollo$^1$, Andrei Mihai Albu$^1$, Alessio Burrello$^1$, Daniele Jahier Pagliari$^1$, \\ Cristian Tesconi$^2$, Loris Panaro$^2$, Dario Soldi$^2$, Fabio Autieri$^2$, Sara Vinco$^1$}
\IEEEauthorblockA{ 
$^1$ Politecnico di Torino, Italy - $^2$ Dumarey Group, Italy}
\IEEEauthorblockA{Emails: name.surname@polito.it}
}

\maketitle
\copyrightnotice
\begin{abstract}
The recent advancements of the automotive sector %
demand robust co-simulation methodologies that enable early validation and seamless integration across hardware and software domains. However, the lack of standardized interfaces and the dominance of proprietary simulation platforms pose significant challenges to collaboration, scalability, and IP protection. To address these limitations, this paper presents an approach for automatically wrapping SystemC models by %
using the Functional Mock-up Interface (FMI) standard. This method combines the modeling accuracy and fast time-to-market of SystemC with the interoperability and encapsulation benefits of FMI, enabling secure and portable integration of embedded components into co-simulation workflows. We validate the proposed methodology on real-world case studies, %
demonstrating its effectiveness with complex 
designs.
\end{abstract}

\begin{IEEEkeywords}
SystemC, Functional Mockup Unit, Functional Mockup Interface, Software-defined Vehicle
\end{IEEEkeywords}
\section{Introduction}
\label{sec:introduction}

The automotive industry is undergoing a significant transformation,
driven by the rapid development of electric vehicles
(EVs), autonomous driving systems, and increasingly
sophisticated driver assistance features \cite{9046805}. These changes impact the complexity of vehicle architectures, and require
robust design methodologies that allow for early
validation, rapid iteration, and reliable system integration \cite{Crespi2023}.

In parallel, the industry is embracing the Software-Defined Vehicle (SDV) paradigm, where software functionality plays a central role in vehicle capabilities \cite{Liu2022,10136910}. %
This transformation requires %
comprehensive system validation strategies \cite{zhou2016hardwaresoftwarecodesignautomotive, Schulze2022}. 
In this context, co-simulation is crucial, as it enables: 
\begin{itemize}
    \item the joint simulation of different subsystems, when the physical prototypes are not yet available; 
    \item the analysis of complex systems under realistic operating conditions, with each subsystem being simulated in its native simulation environment; 
    \item the synchronized simulation of software modules alongside physical components and control hardware, ensuring that embedded software meets performance and safety requirements under diverse operating conditions \cite{676692}.
\end{itemize}

To ensure the effectiveness of co-simulation in large, multidisciplinary projects, standardization of simulation interfaces is crucial, as %
a mean to promote tool interoperability, improve scalability, and enhance maintainability throughout the development lifecycle. Additionally, the increasing role of collaborative development among original equipment manufacturers, %
suppliers, and software vendors, implies a growing need to protect Intellectual Property (IP), as sharing detailed source models between parties often poses confidentiality and competitive risks. Therefore, simulation approaches that encapsulate models, while allowing at the same time their simulation, %
are highly desirable.

Many existing co-simulation solutions are proprietary platforms %
\cite{simulink, etas-cosym}. These platforms often come with limitations regarding licensing, extensibility, and interoperability. %
To address these limitations, there is increasing momentum toward open-source solutions that offer transparency, customizability, and a lower entry barrier.  

In this context, the Functional Mock-up Interface (FMI) standard has gained wide adoption \cite{FMI_reference}. %
FMI defines a tool-agnostic interface for exchanging and co-simulating dynamic systems. The standard allows models to be exported as packaged containers encapsulating compiled code, metadata, and interface specifications. 
This makes FMI a promising solution for scenarios that require standardized communication and safe IP model sharing. FMI currently supports many languages and frameworks in the context of automotive design flows, including Modelica, Simulink, and Dymola \cite{10393858}. 

The SDV paradigm shift is on the other hand increasing  the relevance of hardware/software co-design methodologies and languages. In particular, SystemC has established itself as a leading modeling language for embedded systems and hardware architectures \cite{SystemC_Reference, Muttillo2023}, as its event-driven simulation capabilities and high-level abstraction make it well-suited for modeling complex digital systems and their timing behavior. SystemC models are often used to describe components such as ECUs, sensors, and controllers, which are central to modern vehicle platforms \cite{Niimi2012, Ishihara2012}. However, SystemC lacks native support for standardized co-simulation interfaces, making its integration into broader system simulations challenging. Some research efforts tried to integrate SystemC in the FMI flows, but with limitations in terms of support for synchronization, interrupt modeling, and automation \cite{Using_FMI_components_in_discrete_event_systems,Kim_related,Ravi_related,8632395,Centomo_paper,Bucs_et_al,7306083,Safar2018,10.1145/3339985.3358488}.

To bridge this gap, we propose an approach that \emph{automatically wraps SystemC models by using the FMI standard}. This hybrid solution combines the modeling precision of SystemC with the interoperability and encapsulation benefits of FMI. By doing so, we aim at creating a portable and secure modeling interface that can be easily integrated into larger co-simulation environments, still supporting IP protection and providing a standardized interface for simulation.

The paper is structured as follows: Section \ref{sec:background} provides the necessary background and state of the art. Section \ref{sec:methodology} focuses on the methodology and its automation, that is then applied to case studies in Section \ref{sec:results}. Finally, Section \ref{sec:conclusions} draws our concluding remarks.

\section{Background and Related Works}
\label{sec:background}

\subsection{Functional Mock-up Interface}
\label{sec:fmi}
FMI is a standardized, open interface that facilitates the exchange and co-simulation of dynamic models across different simulation tools \cite{FMI_reference}. It enables the encapsulation of models into self-contained components called Functional Mock-up Units (FMUs), which can be easily shared, reused, and integrated across various platforms without exposing proprietary information or requiring access to the original source code and modeling environment. 

FMUs execute according to three different solutions: Model Exchange (all FMUs are described as systems of equations, solved by a shared solver), Scheduled Execution (FMUs share a common external scheduling algorithm), and Co-Simulation (each FMU incorporates its own solver and autonomously progresses its internal state according to specified communication intervals). In the automotive context, given the heterogeneity of the domains to be covered, Co-Simulation is the most frequent choice, and is thus the target of this work. 

An FMU is structured as a compressed archive (\texttt{.fmu} file) containing several key elements: 
\begin{itemize}
    \item An XML-based model description file (\texttt{modelDescription.xml}) that defines variables, parameters, input/output signals, data types, and model structure.
    \item Platform-specific binary files (shared libraries, e.g., \texttt{.dll}, \texttt{.so}) that implement the simulation logic.
    \item Optional resources such as documentation, initialization files, lookup tables, or graphics.
\end{itemize}

The FMI standard defines an extensive set of function calls to support consistent and flexible operation across simulation tools. Full specifications are available here \cite{FMI_reference}; for simplicity, only a subset of these functions is listed below:
\begin{itemize}
    \item \texttt{fmi3Instantiate}: creates a new instance of an FMU; 
    \item \texttt{fmi3SetXXX}: function to set the value of a variable, where \texttt{XXX} is one of the FMI data types (e.g., \texttt{Int8}, \texttt{Int32}, \texttt{Int64}, \texttt{Uint}, \texttt{String}, \texttt{Boolean}); %
    \item \texttt{fmi3DoStep}: advances simulation by a specific time interval;
    \item \texttt{fmi3GetXXX}: function to retrieve the value of a variable, where \texttt{XXX} is one of the FMI data types; 
    \item \texttt{fmi3FreeInstance}: releases resources allocated to an FMU instance when no longer needed. 
\end{itemize}

\subsection{SystemC}
SystemC is a powerful C++ class library that extends standard C++ with hardware modeling capabilities, enabling system-level design and verification of complex digital systems \cite{SystemC_Reference}. It provides a unified framework where both hardware and software components can be described, simulated, and verified together at various level of abstraction, from high-level functional models to detailed Register-Transfer Level (RTL) implementations. 
As its core, SystemC consists of: 
\begin{itemize}
    \item A discrete time, event-based simulation kernel that schedules and executes processes on the occurrence of synchronizations, time notifications or signal value changes, with specific functions to start (\texttt{sc\_start()}) and pause (\texttt{sc\_pause()}) simulation and to dynamically generate new processes whenever needed (\texttt{sc\_spawn()});  
    \item A potentially hierarchical organization of modules (\texttt{SC\_MODULE}), which encapsulate functionality in processes, and communicate via ports and channels. %
    \item Primitive channels and interfaces for communication between components.
    \item Hardware-oriented language constructs, %
    including specific data types, such as bit vectors and logic values (e.g., \texttt{sc\_bit}, \texttt{sc\_bv}, \texttt{sc\_logic}, \texttt{sc\_lv}). 
\end{itemize}

This paper focuses specifically on SystemC RTL, where %
simulation relies on cycle-accurate timing models to reflect hardware clock behavior, and on a detailed signal semantics to capture transitions and propagation delays through combinatorial logic. %
As a result, SystemC offers a versatile and efficient platform for modeling and simulating hardware systems with high fidelity, making it well-suited for both early-stage design exploration and detailed implementation.

\subsection{Related Works}
\label{sec:related-works}
Ever since its first release in 2010, the FMI standard has garnered significant attention from both academia and the automotive industry \cite{fmi_original_paper}, that has led to its widespread adoption across various simulation environments.

In \cite{Using_FMI_components_in_discrete_event_systems}, the authors propose one of the earliest attempts to integrate FMI for Model Exchange into discrete event systems, where FMI components generated with OpenModelica are embedded within the discrete event domain of Ptolemy II.
More recently, \cite{Kim_related} presents an FMI 2.0-based approach for integrating CAN bus simulation into Simulink-based vehicle simulations, including %
realistic CAN communication between various vehicle subsystems, message transmission timing and priority-based arbitration. %
In \cite{Ravi_related}, a modular co-simulation architecture for timing-aware Software-in-the-Loop (SiL) simulation of automotive applications using the FMI 3.0 standard is proposed. Their approach couples timing simulation with functional simulation, allowing for early evaluation of software behavior on target hardware. The architecture consists of a timing simulator, virtual ECU, communication point service, and mode service, each implemented as a Co-Simulation FMU. The case study demonstrates how timing-aware simulations can reveal functional behavior deviations that are not captured by traditional SiL testing, potentially reducing reliance on Hardware-in-the-Loop validation.

\begin{table}[!th]
    \centering
    \caption{Related works}
    \label{tab:related_works}
    \resizebox{\columnwidth}{!}{%
    \begin{tabular}{|c|c|c|c|}
    \cline{2-4}
    \multicolumn{1}{c|}{} & FMI & SYSTEMC & METHODOLOGY \\     
    \multicolumn{1}{c|}{} & VERSION & SUPPORT & AND TOOLS \\ \hline\hline
    \cite{Using_FMI_components_in_discrete_event_systems} & 2.0 & \no & OpenModelica and Ptolemy II required  \\ \hline
    \cite{Kim_related} & 3.0 & \no & Custom for CAN bus \\ \hline
    \cite{Ravi_related} & 3.0 & \no & No tools required \\ \hline \hline 
    \cite{8632395} & 1.0 & $\approx$ & DEVS and Matlab required \\ \hline
    \cite{Centomo_paper} & 2.0 & \yes & No interrupts \\ \hline
    \cite{Bucs_et_al} & 2.0 & \yes & Platform Architect required \\ \hline
    \cite{7306083} & 2.0 & \yes & Overhead in the integration  \\ \hline 
    \cite{Safar2018} & 2.0 & \yes & Overhead in the integration   \\\hline
    \cite{10.1145/3339985.3358488} & 2.0 & \yes & Requires VPSim \\ \hline \hline
    \rowcolor{LightCyan}
    & (1.0) & &  No tools required \\
    \rowcolor{LightCyan} Our & (2.0) & \yes & Simple Integration \\
    \rowcolor{LightCyan} & 3.0 & & Full Interrupt support \\
    \hline
    \end{tabular}}
\end{table}
However, none of the approaches mentioned above address the integration of SystemC, or any other hardware description language, within the FMI standard. One attempt to bridge this gap is presented in \cite{8632395}, where the authors use FMI 1.0 to co-simulate a hardware model specified at the RTL level using the DEVS Suite Simulator with a software model defined as a MATLAB script. %
\cite{Bucs_et_al} tackled the integration challenge by proposing a method to incorporate SystemC-based virtual prototypes into heterogeneous, multi-domain automotive simulations via the FMI standard. However, their solution relies heavily on Platform Architect, a commercial tool from Synopsys for assembling virtual prototypes \cite{platform-architect}.
In \cite{7306083}, a methodology is proposed for integrating both SystemC and SystemC/AMS models into FMUs using FMI 2.0. While the method is general, it requires the addition of external functions outside of the FMI specification to manage SystemC execution.
\cite{Centomo_paper} introduced a bridge between SystemC and FMI 2.0 to enable exporting SystemC models as FMUs. While effective, the authors identified limitations in FMI’s native support for event-driven communication and data types. Additionally, interrupt handling was not considered in their work.
In \cite{Safar2018}, a generic and configurable FMI master is developed to facilitate communication between SystemC/TLM-based virtual platforms and FMUs from various tools. This approach supports multiple FMU slaves but necessitates modifying the SystemC design, as the FMI master is instantiated within the top-level module encapsulating the SoC model.
Lastly, \cite{10.1145/3339985.3358488} presents a general and scalable methodology for co-simulating Cyber-Physical Systems (CPS) using FMI. However, the method is closely tied to the VPSim environment for virtual prototyping \cite{vpsim}, thus limiting its portability.

An overview of the related work discussed above is provided in Table \ref{tab:related_works}. In summary, existing methods either depend on proprietary tools, lack seamless integration with the FMI standard, or require modifications to the SystemC design. Our proposed approach addresses these limitations by offering a fully open-source, simple, and extensible solution to integrate any SystemC design into an FMU.

\section{Methodology}
\label{sec:methodology}

\begin{figure}[ht]
    \centering
    \includegraphics[width=1\linewidth]{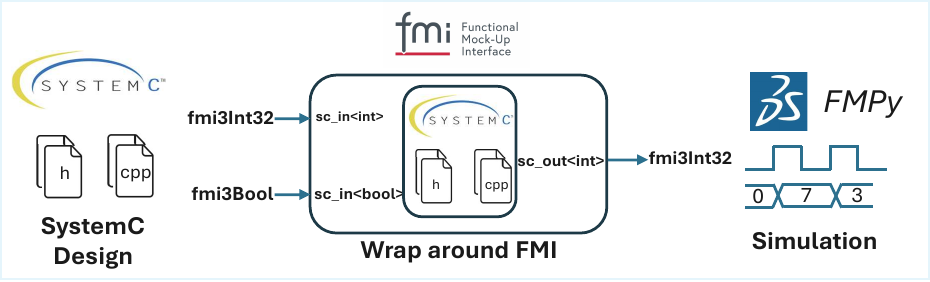}
    \caption{Methodology Overview in three steps: input SystemC Design, generation of the FMI wrapper and Simulation driven by a FMI master}
    \label{fig:methods-overview}
\end{figure}

This section provides a detailed explanation of the methodology, starting with a high-level overview (Figure \ref{fig:methods-overview}) and progressively diving into each specific detail. Finally, it presents the automation framework, which integrates all methodological aspects in a plug-and-play fashion, making the use of the tool almost transparent to the end user.

\subsection{High Level Overview}
The methodology proposed in this paper is a three-step flow, as summarized in Figure~\ref{fig:methods-overview}. %
From left to right, the first step is the selection of the {SystemC design}, that is the entry point of the flow. %
In the second step, the SystemC design is encapsulated in a FMI wrapper, that %
ensures a standard co-simulation interface with no internal modifications or adaptations of the design itself. This phase allows to map input and output ports to corresponding FMI data types, to implement the necessary FMI-compliant functions, and to generate the resulting \texttt{.fmu} file.
Finally, the flow concludes with the {simulation} of the design, performed by using a master that invokes the FMI functions to control the evolution of simulation. The master can be either implemented with an open library, e.g., FMPy \cite{FMPy_library}, or within proprietary tools, thus extending the applicability of the methodology to commercial environments.

\subsection{SystemC Design Pre-Processing}
The entry point of the methodology is the initial SystemC design, that must be wrappped as a FMU. The proposed methodology does not impose any limitation on the chosen design and supports {all designs written in SystemC at the RTL level}. The only assumption is that the design contains a top level entity with input/output ports to be mapped onto FMU variables, and that the source code of the interface is accessible to extrapolate the list of such ports. Such design may be hierarchical, i.e., contain a number of SystemC modules. 
Given that the proposed approach %
does not require source code modifications, %
any SystemC version and any design for which the source code is accessible can be directly integrated.

The top-level entity of the SystemC design is parsed to extrapolate information necessary to %
define the 
\texttt{ModelDescription.xml} file \cite{FMI_reference}, that %
contains all necessary metadata for the FMU (e.g., the FMI version and the model name), as well as a complete description of the SystemC interface. %
The SystemC design is thus parsed, to extrapolate the complete list of ports, complete of their name, direction (input/output), type. 

\subsection{FMI wrapper construction}
\begin{table}
    \centering
    \caption{Mapping of the ports between SystemC and FMI}
    \begin{tabular}{|c|c|}
        \hline
         {SYSTEMC DATA TYPE} & {FMI DATA TYPE}  \\
         \hline%
         \texttt{sc\_logic} & \texttt{fmi3Bool} \\
         \hline%
         \texttt{sc\_bv$<$N$>$} & \texttt{fmi3Binary} \\
         \hline%
         \texttt{sc\_int$<$1 .. 8$>$} & \texttt{fmi3Int8} \\
         \texttt{sc\_int$<$9 .. 16$>$} & \texttt{fmi3Int16} \\
         \texttt{sc\_int$<$17 .. 32$>$} & \texttt{fmi3Int32} \\
         \texttt{sc\_int$<$33 .. 64$>$} & \texttt{fmi3Int64} \\
         \hline%
         \texttt{sc\_uint$<$1 .. 8$>$} & \texttt{fmi3UInt8} \\
         \texttt{sc\_uint$<$9 .. 16$>$} & \texttt{fmi3UInt16}\\
         \texttt{sc\_uint$<$17 .. 32$>$} & \texttt{fmi3UInt32} \\
         \texttt{sc\_uint$<$33 .. 64$>$} & \texttt{fmi3UInt64} \\
         \hline%
         \texttt{sc\_float} & \texttt{fmi3Float32} \\
         \hline%
         \texttt{sc\_double} & \texttt{fmi3Float64} \\
         \hline 
    \end{tabular}
    \label{tab:systemc-fmi-ports}
\end{table}

The next step consists of wrapping the SystemC design with the necessary FMI 3.0 constructs. This wrapping is implemented as a 
container that includes both SystemC and FMI components, to bridge the two domains, necessary to allow and control SystemC simulation, and to make %
the values of SystemC ports accessible from the FMI functions, to set the correct inputs and gather the corresponding outputs. 
This is achieved with the declaration of a \texttt{struct} containing a reference to an instance of the SystemC top level entity (allocated at run time), plus one FMI variable and one SystemC signal for each SystemC input and output port  
(middle diagram of Figure~\ref{fig:methods-overview}): 
\begin{itemize}
\item the signals will be used to bind the ports, as all top-level entity ports must be bound to either ports or signals to allow successful SystemC simulation. The signals have the same type as the original SystemC port, and their name is the same as the port, with a \texttt{s\_} suffix; 
\item the variables will be used to store the corresponding values in an FMI-compatible format. The name of the variables is the same as the original SystemC port. Each port type is mapped onto an FMI type, as summarized in Table~\ref{tab:systemc-fmi-ports}. %
In addition to SystemC-specific types, all standard C++ data types are also supported and mapped accordingly (e.g., \texttt{int} is mapped to \texttt{fmi3Int32}, \texttt{float} to \texttt{fmi3Float32}, etc.). This mapping is designed to be extensible: any new types introduced in future FMI versions will be coherently mapped to the closest matching C++/SystemC data types.
\end{itemize}

The next step involves implementing the prototype functions required by the FMI standard. These functions define the core interactions between the simulation environment and the SystemC design: %
\begin{itemize}
    \item the \texttt{fmi3InstantiateCoSimulation} function instantiates the SystemC top level design and binds the SystemC signals to the design ports. %
    At this point, any further initialization can be applied (e.g., to set default values);   

\item the \texttt{fmi3SetXXX} and \texttt{fmi3GetXXX} functions are then used to map values of the FMI variables onto the SystemC signals and vice versa. This ensures data passing between the FMI domain and the SystemC domain, to give the correct inputs throughout the simulation and gather the corresponding results;

\item the \texttt{fmi3DoStep} is used to manage the progression of SystemC simulation, as will be detailed in the next section; 

\item the \texttt{fmi3FreeInstance} function frees any dynamic allocated memory, and invokes the destructor of the SystemC top level entity. 
\end{itemize}

After the complete set of prototype functions has been developed and verified, the design becomes ready for compilation. According to the FMI standard, a platform-specific compilation is expected to produce an executable library, e.g., a Dynamic Link Library (\texttt{.dll}) for Windows or a  shared object library (\texttt{.so}) for Linux systems. %

\subsection{FMI-controlled SystemC Simulation}
\label{sec:dostepalternatives}
The activation of SystemC simulation must respect both the SystemC execution semantics (without requiring modifications to the design and/or to the simulation kernel) and the FMI simulation structure. At the same time, %
timing plays a crucial role in the RTL simulation paradigm, and it must thus be %
properly synchronized between the two environments. %

\begin{figure}[ht]
    \centering
    \includegraphics[width=1\linewidth]{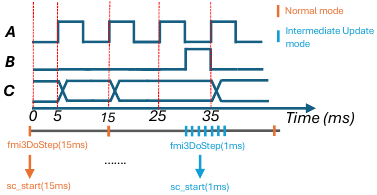}
    \caption{Step (orange) and intermediate update (blue) modes.}
    \label{fig:normal-intermediate-modes}
\end{figure}

The FMI standard allows three main types of execution:

\paragraph{Step mode} The first mode activates the FMUs at fixed time steps. In this scenario, the \texttt{ModelDescription.xml} file contains a \texttt{CommunicationStepSize} parameter, that conveys the size of the fixed time step to be used at any FMU invocation. 
To synchronize the SystemC simulation with this mechanism, an invocation of the \texttt{fmi3DoStep} internally triggers SystemC simulation with an invocation of the \texttt{sc\_start()} for a duration corresponding to the specified \texttt{CommunicationStepSize}. 
An example of this mechanism is visible in Figure \ref{fig:normal-intermediate-modes}, where each \texttt{fmi3DoStep(15ms)} triggers the \texttt{sc\_start(15ms)}.
This straightforward encapsulation ensures synchronization between the SystemC simulation and the FMI co-simulation environment without requiring any modification to the SystemC simulation kernel.

\paragraph{Intermediate update}\label{sec:intermediate} This second mechanism %
is provided by the FMI standard to enable the exchange of input and output values between communication points. This is particularly useful in scenarios where a finer simulation granularity is required (e.g., in case of physical simulations). When the FMU \texttt{IntermediateUpdateMode} is detected, the framework dynamically reduces the \texttt{CommunicationStepSize} used by the \texttt{fmi3DoStep}, to advance SystemC simulation time with finer temporal resolution. A visualization of this mechanism is represented in Figure \ref{fig:normal-intermediate-modes}, where at a certain point of the simulation, the \texttt{CommunicationStepSize} becomes of 1ms (blue dashes), exploiting a finer grain execution.

\begin{figure}[ht]
    \centering
    \includegraphics[width=.9\linewidth]{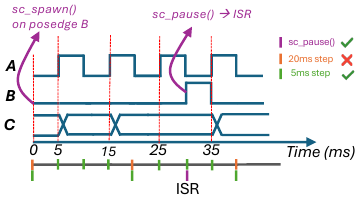}
    \caption{Event-based/Interrupt-driven mode with the two proposed solutions.}
    \label{fig:interrupts}
\end{figure}

\paragraph{Interrupt management}\label{sec:interrupt}
The final and arguably most critical timing mechanism is the {event-based or interrupt-driven execution}. Handling interrupts requires a more sophisticated management approach, as an interrupt typically triggers the execution of an Interrupt Service Routine (ISR). Two solutions have been developed to address this need, differing mainly in their invasiveness and complexity. Both solutions are illustrated in Figure \ref{fig:interrupts}, where $B$ is the signal treated as interrupt.

The first approach, which is capable of handling asynchronous interrupts, requires a minimal adaptation of the SystemC wrapper around FMI, although it does not necessitate any modifications to the original SystemC design. %
The core idea is to generate a dynamic SystemC process (with the %
\texttt{sc\_spawn()} primitive), used to detect when a specified condition is met (e.g., the rising edge of the signal interrupt). Upon detecting such condition, the system enters the FMI \texttt{EventMode} \cite{FMI_event_mode} and calls the \texttt{sc\_pause()} primitive, that pauses the SystemC simulation and the execution of the \texttt{fmi3DoStep}. The ISR is then executed, after which the system returns to FMI \texttt{StepMode} \cite{FMI_step_mode}, thus resuming normal simulation until the next interrupt occurs. This allows handling asynchronous interrupts with high precision. 

The second approach is engineered to avoid any modifications to the internal simulation code, thus maximizing portability and simplicity. However, it is limited to managing synchronous interrupts and its accuracy depends on the simulation step size. In this case, one or more signals are selected for monitoring, and at the end of each \texttt{fmi3DoStep}, their values are checked against the desired interrupt condition. If the condition is met, the ISR is executed. The limitation of this solution is clearly that its ability to detect interrupts is directly tied to the simulation’s temporal resolution. In Figure \ref{fig:interrupts},  the finer step size ($5ms$, green) allows for successfully detection of the interrupt, while the larger step size ($20ms$, orange), misses the interrupt. A correct sizing of the 
\texttt{CommunicationStepSize} size parameter is thus crucial. On the other hand, %
this solution offers a good trade-off between complexity and simulation accuracy when a sufficiently fine simulation step is adopted. 
This trade-off is an intentional design choice aimed at maximizing portability and simplicity, specifically to avoid any modifications to the SystemC wrapper. While this may come at the cost of reduced accuracy, it enables a non-invasive integration strategy that remains broadly compatible and easier to adopt.

\subsection{Simulation using FMPy}

The final stage of the proposed workflow consists of simulating the generated FMU (rightmost block of Figure~\ref{fig:methods-overview}). For this purpose, the FMPy Python library was selected. FMPy simplifies the simulation process by exposing high-level APIs that internally call the standard FMI functions, facilitating seamless execution \cite{FMPy_library}. The library offers full support for all functions defined in the FMI 3.0 specification and is compatible with each of the execution modes described in the previous sections. However, it is important to note that the generated FMU can be integrated in any FMU-based environment. 

\subsection{Automation Framework}
\label{sec:automation}
\begin{figure*}
    \centering
    \includegraphics[width=1\linewidth]{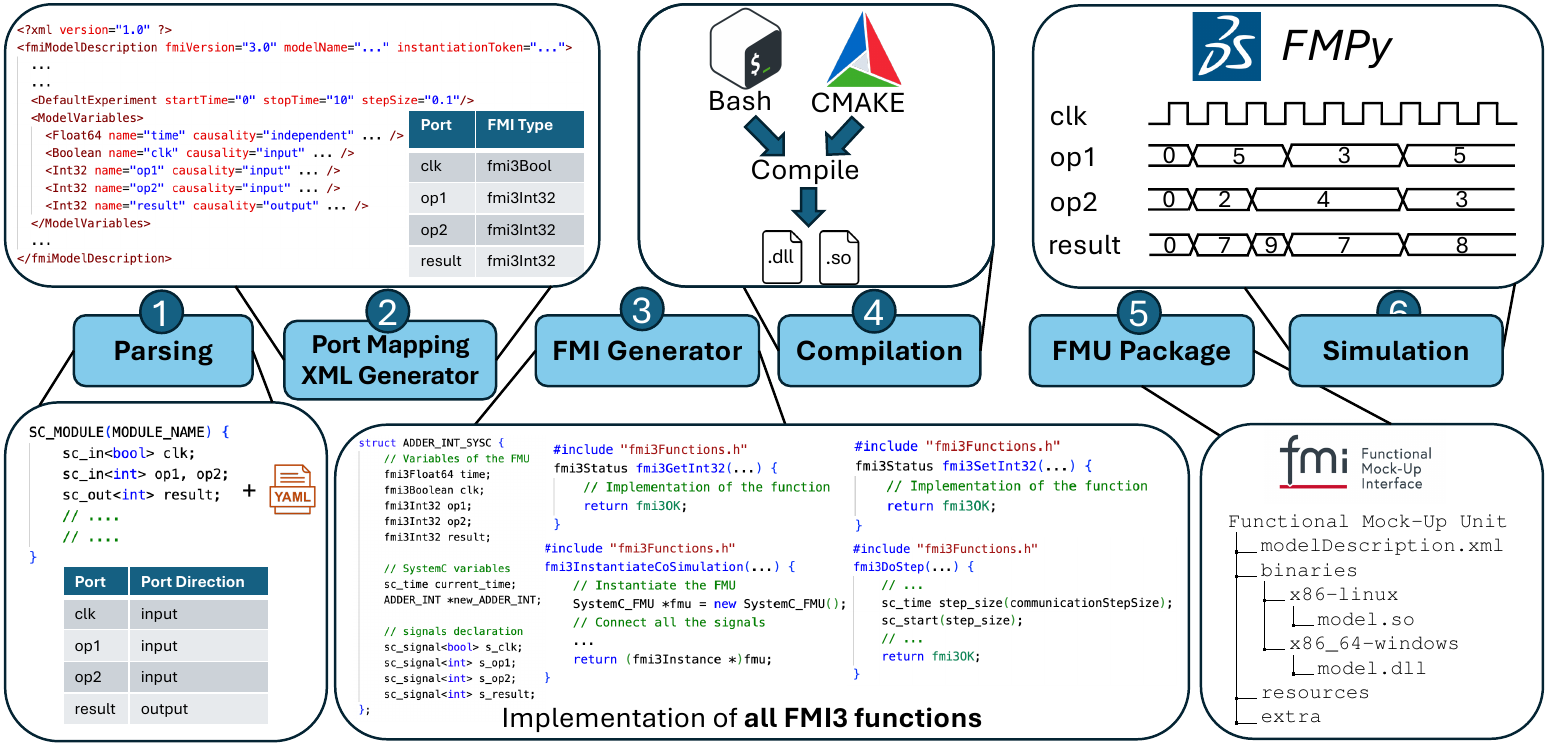}
    \caption{Automation Framework exemplified on the ALU case study.}
    \label{fig:automation}\vspace{-0.5cm}
\end{figure*}

The previous sections have primarily focused on the methodology and internal mechanisms of the proposed framework, with limited attention given to implementation-specific details. This separation is intentional, as all previously described processes have been automated to maximize usability and accessibility. To this end, a lightweight yet extensible automation framework has been developed, aimed at streamlining FMU generation and simulation with minimal user intervention. An overview is reported in Figure \ref{fig:automation}.

The automation process takes as input two elements: the SystemC design and a configuration file written in YAML format. The YAML file serves multiple roles, including defining the paths to various SystemC source files and specifying configuration parameters for FMU generation. It also simplifies the definition of the FMU’s XML metadata by abstracting the verbosity of the \texttt{ModelDescription.xml} into a minimal set of user-defined fields.
Additional configuration parameters include simulation-related attributes such as step size and total simulation duration. 

Once configured, the automation is triggered by executing a Python script that carries out the following steps:
\begin{enumerate}
    \item \textbf{Parsing} - Analyzes the SystemC design to identify its input and output ports.
    \item \textbf{Model Description Generation} - Creates the \texttt{ModelDescription.xml} file using both the parsed port information and user-provided configuration.
    \item \textbf{FMI Function Implementation} -  Generates all required FMI functions, incorporating the appropriate port mappings described in Section~\ref{tab:systemc-fmi-ports}.
    \item \textbf{Compilation} - Compiles both the SystemC design and the generated FMI glue code using either a Bash script or CMake, depending on user preference.
    \item \textbf{FMU Packaging} - Packages the compiled binaries into a platform-specific FMU, producing a \texttt{.dll} for Windows or a \texttt{.so} for Linux, thus supporting both Operating Systems.
    \item \textbf{Simulation Execution} - Launches a simulation run using the newly created FMU.
\end{enumerate}

This automation pipeline enables a nearly hands-free workflow, significantly reducing the barrier to entry for users who need to encapsulate SystemC designs into FMUs for co-simulation purposes. The framework is available at \texttt{https://github.com/eml-eda/systemc-fmi}. %

\section{Experimental Results}
\label{sec:results}

To evaluate the effectiveness of the proposed integration method, we conducted an extensive series of experiments using a set of heterogeneous SystemC models, including custom-developed examples, industrial case studies by partner companies, and publicly available models sourced from GitHub. 
Each case study underwent the integration process outlined in Figure \ref{fig:automation}, %
after which we carried out several performance assessments. Experiments were conducted on Ubuntu 24.04 LTS running inside WSL, on a host machine equipped with an Intel i7-12700H processor and 16GB of RAM.
Specifically, we measured and reported in Table \ref{tab:experimental-results}: 
\begin{itemize}
\item \emph{Model complexity}, evaluated based on the number of: ports, processes, lines of code (LoC), and of hierarchical modules, that should give a measure of the size and complexity of each case study;
\item \emph{Total simulation time}, comparing native SystemC execution to simulation as FMU controlled via the FMPy library. The profiling infrastructure for the native SystemC version was developed in C++ ($s$ columns), whereas the FMU was assessed using the Python-based co-simulation framework using libraries such as \texttt{psutil} \cite{psutil} and is reported as slow down factor, i.e., ratio between the FMU-based and native SystemC execution times ($\times$ columns). Each experiment was executed five times, and the average of the results was computed to enhance the reliability of the collected data. Both the encapsulated FMU model and the native SystemC implementation were evaluated under varying simulation durations, ranging from 100 ms to 10s, and the step size of FMU execution was set to 1 ms for all experiments;
\item \emph{Memory overhead}, reported as SystemC memory request ($MB$ columns) and FMI-induced overhead ($\times$ columns).
\end{itemize}

\begin{table*}[]
\caption{Experimental results}
\label{tab:experimental-results}
\resizebox{\textwidth}{!}{%
\begin{tabular}{c|c|c|c|c||r|r|r|r|r|r||r|r|r|r|r|r|}
\cline{2-17}
 &
  \multicolumn{4}{c||}{{DESIGN CHARACTERISTICS}} &
  \multicolumn{6}{c||}{{EXECUTION TIME}} &
  \multicolumn{6}{c|}{{MEMORY}} \\ \cline{2-17} 
 &
  \multicolumn{1}{c|}{{{PORTS}}} &
  \multicolumn{1}{c|}{{{PROCESSES}}} &
  \multicolumn{1}{c|}{{{MODULES}}} &
  \multicolumn{1}{c||}{{{LoC}}} &
  \multicolumn{2}{c|}{{100 ms}} &
  \multicolumn{2}{c|}{{1 s}} &
  \multicolumn{2}{c||}{{10 s}} &
  \multicolumn{2}{c|}{{100 ms}} &
  \multicolumn{2}{c|}{{1 s}} &
  \multicolumn{2}{c|}{{10 s}} \\ \cline{6-17} 
 &
  \multicolumn{1}{c|}{(\#)} &
  \multicolumn{1}{c|}{(\#)} &
  \multicolumn{1}{c|}{(\#)} &
  \multicolumn{1}{c||}{(\#)} &
  \multicolumn{1}{c|}{{s}} &
  \multicolumn{1}{c|}{{$\times$}} &
  \multicolumn{1}{c|}{{s}} &
  \multicolumn{1}{c|}{{$\times$}} &
  \multicolumn{1}{c|}{{s}} &
  \multicolumn{1}{c||}{$\times$} &
  \multicolumn{1}{c|}{{MB}} &
  \multicolumn{1}{c|}{{$\times$}} &
  \multicolumn{1}{c|}{{MB}} &
  \multicolumn{1}{c|}{{$\times$}} &
  \multicolumn{1}{c|}{{MB}} &
  \multicolumn{1}{c|}{$\times$} \\ \hline\hline
\multicolumn{1}{|l|}{{ALU}} &
  4 & 1 & 1 & 317 & 0.003 & 15.98 &
  0.012 &
  15.16 &
  1.421 &
  {12.08} &
  4.650 &
  24.53 &
  4.620 &
  25.03 &
  4.650 &
  {27.79} \\ %
  \hline
\multicolumn{1}{|l|}{{CRC}} &
  12 &
  15 &
  1 &
  {2,467} &
  0.021 &
  4.07 &
  0.082 &
  2.30 &
  0.642 &
  {1.84} &
  5.530 &
  19.91 &
  5.560 &
  19.92 &
  5.510 &
  {20.93} \\ %
  \hline
\multicolumn{1}{|l|}{{I2C}} &
  10 &
  11 &
  2 &
  {1,085} &
  0.022 &
  8.95 & 
  0.134 &
  9.53 &
  2.362 &
  {6.28} &
  5.547 &
  18.75 &
  5.652 &
  18.76 &
  7.336 &
  {17.98} \\ %
  \hline

\multicolumn{1}{|l|}{{RISC}} &
  {32} &
  {8} &
  {6} &
  731 &
  {0.013} &
  {11.78} &
  {0.029} &
  {12.34} &
  {0.138} &
  14.87 &
  {5.280} &
  {21.57} &
  {5.332} &
  {21.61} &
  {5.036} &
  21.78 \\ \hline

  \multicolumn{1}{|l|}{Delta Sigma} &
  4 &
  1 &
  1 &
  {354} &
  0.033 &
  14.85 &
  0.330 &
  4.38 &
  3.283 &
  {3.61} &
  4.430 &
  23.12 &
  4.434 &
  23.21 &
  5.270 &
  {20.30} \\ %
  \hline
\end{tabular}%
}
\vspace{0.5mm}
\end{table*}

\subsection{{Arithmetic Logic Unit}}

As an introductory example to illustrate the overall integration workflow in Figure \ref{fig:automation}, we selected a SystemC design of an Arithmetic Logic Unit (ALU) containing a single synchronous process featuring three unsigned integer inputs (two 4-bit operands and a 3-bit opcode with 8 different operations) and a single 4-bit unsigned integer output representing the result. As discussed in Table~\ref{tab:systemc-fmi-ports}, these data types are mapped to \texttt{fmi3UInt8} variables in the \texttt{modelDescription.xml file}, serving as input and output representations in the FMU. 

After encapsulating the SystemC model as a FMU, a comparative evaluation was performed against the native SystemC implementation. %
As expected, the FMU-based simulation introduces additional performance overhead due to the abstraction layers and runtime interfacing inherent in the FMI framework. The slowdown factor, defined as %
the ratio between the FMU-based and native SystemC execution times, goes from approx. $16\times$ with the shortest simulation to $12\times$ for the longest one. This decreasing trend 
highlights the presence of a fixed overhead introduced by the FMI infrastructure, mostly associated with one-time operations such as \texttt{fmi3InstantiateCoSimulation}, \texttt{fmi3SetupExperiment}, and \texttt{fmi3EnterInitializa}- \texttt{tionMode}, executed only during the setup phase. %

In terms of memory usage, the native SystemC simulation consumed approximately 4.6MB in all simulation scenarios, while the encapsulated FMU version increased memory consumption from approx. $24.5\times$ to $28\times$. This rise in memory usage can be attributed to the FMI runtime, that requires operations like data translation and conversion. %
While the memory overhead appears high, especially for such a simple model, it is important to recognize that such cost stems mainly from the flexible, standardized infrastructure defined by the FMI specification, rather than the inherent complexity of the model itself, and by the adoption of a Python-based master, inherently less efficient than a full C++-based implementation. Crucially, the memory usage remains stable across runs and scales predictably, making it acceptable in workflows where modularity, tool interoperability, or platform independence is prioritized over minimal resource consumption.

\subsection{{Cyclic Redundancy Check (CRC)}}
\begin{figure*}
    \centering  \vspace{-0.5cm}
    \includegraphics[width=1\linewidth]{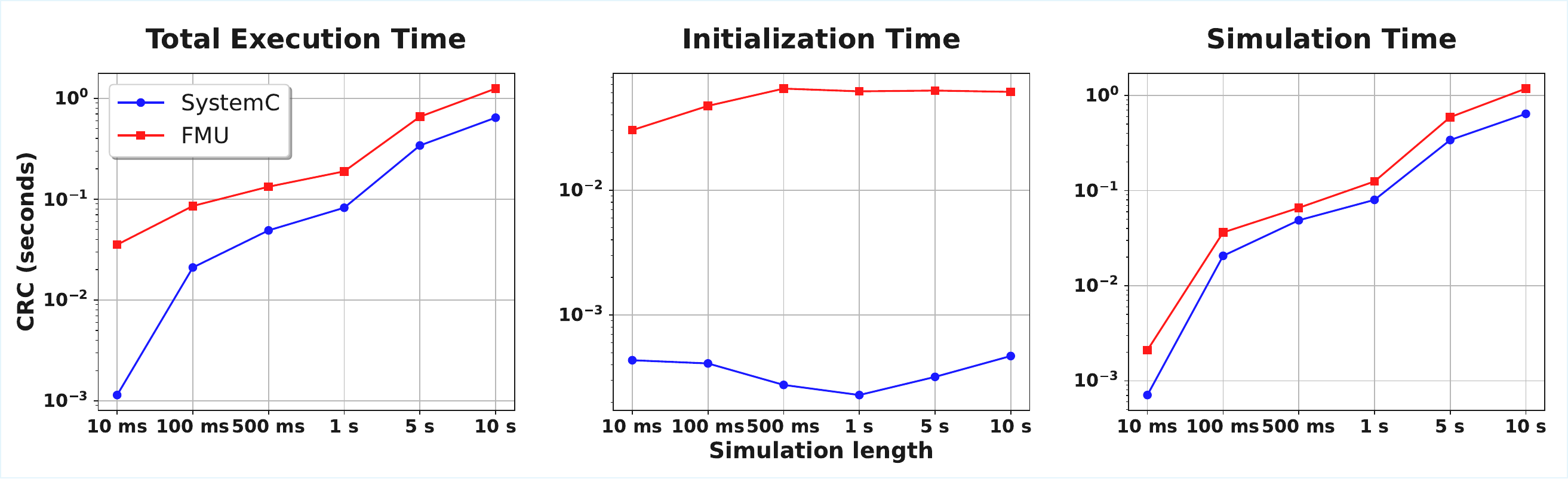}
    \vspace{-0.8cm}
    \caption{CRC simulation performance: SystemC (blue) and FMI-wrapped SystemC (red). Initialization time is almost constant for both SystemC and FMI across all simulation lengths, while the simulation times scales and dominates the total execution time.}
    \label{fig:CRC_performance}\vspace{-0.5cm}
\end{figure*}

The Cyclic Redundancy Check (CRC) case study %
was developed in collaboration with our industry partner for application in automotive verification, where configurable CRC units are critical for error detection in communication protocols such as CAN and Ethernet. This CRC module is a high-performance, configurable hardware model intended for integration into larger System-on-Chip (SoC) verification environments supporting 16 independent CRC contexts, each configurable for either CRC-16 or CRC-32 polynomial operations. Additional features include input/output inversion, byte-swapping, register-based control, error detection capabilities, and support for multi-cycle operation handling. The design incorporates five distinct dataflows and three supporting functions, implemented through a total of 15 processes. 

The results presented in Figure \ref{fig:CRC_performance} highlight the trade-offs in timing and memory performance between the native SystemC model and the encapsulated one. 
The slowdown factor is lower than for the ALU, as the CRC case study includes more computation, that compensates for the FMI-induced overhead. The slowdown factor goes thus from approx. $4\times$ to less than $2\times$, when increasing simulation length. Also in this case there is a fixed overhead introduced by the FMI infrastructure, associated to one-time initialization operations, highlighted by Figure \ref{fig:CRC_performance}: initialization time is almost constant for both SystemC and FMI across all simulation lengths, while the simulation times scales and dominates the total execution time.

In terms of memory consumption, SystemC maintains consistent and minimal memory usage across all test cases of around 5.5MB. Conversely, FMU requires around $20\times$ more memory, in particular in terms of Resident Set Size (RSS, i.e., RAM memory, $20\times$) and of Virtual Memory size (VM, from $10\times$ to almost two orders of magnitude in worst case). %

\subsection{{I2C (Inter Integrated Circuit)}}
As a communication-oriented case study, we adopted an 
open-source I2C bus model, typical of automotive systems \cite{githubGitHubHichemI2C}. %
This case study is hierarchical, i.e., includes two SystemC modules. The master supports core I2C functions ($START/STOP$ conditions, 7-bit addressing, data read/write, and $ACK/NACK$ signaling) via a finite state machine managing protocol phases. The slave responds to address $0x2A$, decoding commands and issuing responses such as data output or write acknowledgments. Both modules manage bidirectional $SDA$ and $SCL$ lines using tri-state control for arbitration, ensuring protocol compliance, clock synchronization, signal integrity. The module supports basic error detection through the $NACK$ signal, set to $1$ for invalid addresses.

The I2C case study is wrapped as a FMU by treating the $NACK$ signal as an interrupt through the \texttt{EventMode} approach (Section \ref{sec:dostepalternatives}.c).
Simulation profiling shows a slowdown factor that scales from $9\times$ for longest simulations to approx. $6\times$ for shortest ones. The higher overhead w.r.t. the former case study can be explained from the need to reinitialize the FMU at each invocation of the \texttt{doStep} function. This overhead increases from 100ms to 1ms long simulations, as it requires a lot of variable set operations, and is compensated only with longer simulations, starting from 10s. 
In this case study, SystemC memory usage increases with time, as an effect of the I2C protocol, but remains lower than 7.5MB. FMI induces an overhead of about $18\times$ ($19\times$ for RSS, $88\times$ for VM, and $7.6\times$ for shared memory). 

\subsection{{RISC Processor}}
As computation- and hierarchy-oriented design, we adopted a SystemC-based Reduced Instruction Set Computing (RISC) processor \cite{githubSystemccpuMaster}. The RISC CPU core is a pipelined processor designed with five stages, that allow the execution of one instruction per %
millisecond. It operates within a memory hierarchy and contains an ALU %
supporting a range of operations including addition, subtraction, logical functions, %
as well as data movement and shift instructions, %
all with integrated flag generation. A dedicated control system governs pipeline operation, managing instruction flow, resolving hazards, and orchestrating access to both memory and registers via control signals. Execution timing is strictly regulated, with each instruction constrained to a one-millisecond window, enforced by a SystemC clock and monitored through second-by-second progress tracking. 

Simulation profiling highlights consistent performance, with a slowdown factor that increases with simulation length, from approx. $12\times$ to $14.5\times$. This is mostly caused by the high number of ports (32), that impose a relatively large overhead at each invocation of the \texttt{fmi3SetXXX} and \texttt{fmi3GetXXX} functions. 
The memory overhead is instead stable across all executions (approx. $21.5\times$). 
These values suggest that the RISC design achieves near-linear scaling w.r.t. simulation duration, implying efficient internal processes and minimal bottlenecks.

\begin{figure}
    \centering
    \includegraphics[width=1\linewidth]{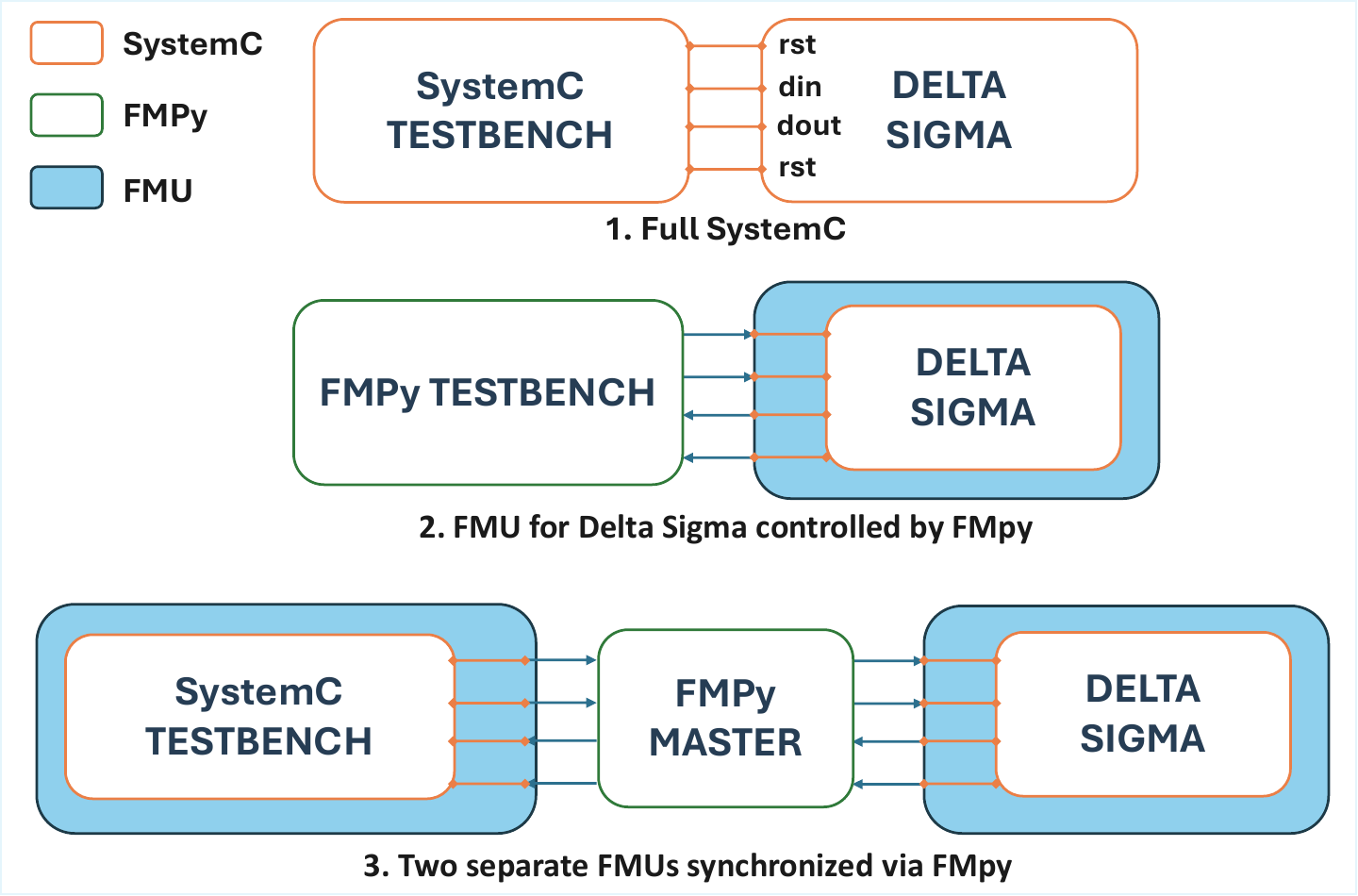}
    \caption{Configurations compared for the Delta Sigma case study.}
    \label{fig:deltasigma}
\end{figure}

\subsection{{Delta Sigma Modulator}}
The last case study is a Delta Sigma modulator provided by an industrial partner. The Delta Sigma converts high-resolution analog input signals into a 1-bit output stream through oversampling and noise shaping. This digital modulator architecture employs n-th cascaded integrator stages that accumulate the difference between the input signal and quantized feedback: at each clock cycle, the system compares the final integrator value against zero to generate the 1-bit output, feeds back its decision to all the n-integrators, and propagates the quantization error through successive integration stages. A synchronous clock generator maintains precise timing with milliseconds resolution, while an active-low reset initializes all integrators to zero. Input is provided by a SystemC testbench, generating a sine wave with variable amplitude and frequency. 

The Delta Sigma case study is wrapped into FMU by allowing also the \texttt{IntermediateUpdate} mode (Section \ref{sec:dostepalternatives}.b), useful to adjust estimation of the FMU output to the variable frequency and amplitude of the generated input. 

In the discussion of this case study, we compare three different configurations, reported in Figure \ref{fig:deltasigma}. The first experiment is in line with the former experiments, and compares the full SystemC system (Figure \ref{fig:deltasigma}.$1$) with the Delta Sigma wrapped as FMU controlled by a FMpy master, feeding the FMU with the same sinusoidal input as the original SystemC testbench (Figure \ref{fig:deltasigma}.$2$). 
Simulation profiling shows a slowdown factor that decreases with the length of simulation, going from approx. $15\times$ to $3.61\times$, as the Delta Sigma is a computation-dominated case study, and compensates well the overhead of initialization and data management induced by FMI. Memory overhead is in the order of $23\times$, similar to the other case studies.

As an additional experiment, we simulated a realistic scenario, where the Delta Sigma is given to a third party as FMU, and the stimuli generation part is a separate FMU interacting via a Python master (Figure \ref{fig:deltasigma}.$3$). To achieve this, we also wrapped the original SystemC testbench as a separate FMU. The memory overhead is similar to the former experiment, as the FMI-induced overhead is the same as when executing the Delta Sigma FMU with a Python master (approx. $23\times$). Simulation times show a very interesting trend: the initial overhead for a 100ms simulation is high (0.033s vs 0.440s, approx. $13\times$), but decreases with the 1s long simulation (0.330s vs 0.750s, $2.27\times$) and becomes almost negligible with the 10s long simulation (3.283s vs 3.679s, $1.12\times$). Two considerations arise from these numbers. The first is that the SystemC sine wave generation is more efficient than the Python one, as configuration $3$ is more efficient than configuration $2$, explored in the former experiment (that for 10s long simulations had a $3.61\times$ overhead).
The second consideration is that SystemC case studies wrapped into FMI by the proposed flow can be effectively given to third parties and integrated in external environments, without disclosing the source code and still allowing good simulation performance.

\section{Conclusions}
\label{sec:conclusions}

In this paper, we proposed an automated framework for integrating SystemC RTL models with the FMI standard, enabling seamless co-simulation between hardware components and software/system-level tools. Our approach facilitates early design space exploration, enhances cross-domain interoperability, and enables third party collaboration.

The framework automatically generates FMUs from SystemC models, exposing standardized interfaces for interaction with FMI-compliant tools. Key advantages include {tool independence}, allowing broad applicability across simulation environments without proprietary restrictions, and {non-invasive integration}, that preserves original SystemC models without manual modifications.

The experimental evaluation demonstrated the framework's efficiency across a range of hardware components with differing complexity levels. The proposed framework is capable of attaining integration and standardisation through the usage of FMI standard, at the price of a performance and memory overhead. %
However, it is important to note that %
the proposed solution allows to wrap complex and heterogeneous case studies, and to allow smooth integration in third party simulations, without disclosing the source code and avoiding any negative impact on simulation accuracy and effectiveness. 

Future work will extend the proposed solution to SystemC TLM, and will analize the integration of the generated SystemC-based FMUs in ISS-based environments (e.g., including QEmu), to prove the integration in the typical %
hardware-software co-design flow for complex embedded systems.

\bibliographystyle{IEEEtran}
\bibliography{references}

\end{document}